%% file: example_paper.tex
\documentclass{article}

\usepackage{microtype}
\usepackage{graphicx}
\usepackage{subfigure}
\usepackage{booktabs} 
\usepackage{pifont}
\usepackage{hyperref}
\usepackage{adjustbox}
\usepackage{multirow}
\usepackage[table, dvipsnames]{xcolor}
\usepackage{tikz}

\usepackage[accepted]{icml2026}

\usepackage{amsmath}
\usepackage{amssymb}
\usepackage{mathtools}
\usepackage{amsthm}

\usepackage[capitalize,noabbrev]{cleveref}

\theoremstyle{plain}

\theoremstyle{definition}

\theoremstyle{remark}

\def\equationautorefname~#1\null{Eq.~(#1)\null}
\newcommand{\aref}[1]{\hyperref[#1]{Appendix~\ref{#1}}}

\usepackage[textsize=tiny]{todonotes}

\input{math_commands}

\icmltitlerunning{Optimal Brain Decomposition}

\begin{document}

\twocolumn[
\icmltitle{Optimal Brain Decomposition for Accurate LLM Low-Rank Approximation}



\icmlsetsymbol{equal}{*}

\begin{icmlauthorlist}
\icmlauthor{Yuhang Li}{yale}
\icmlauthor{Donghyun Lee}{yale,usc}
\icmlauthor{Ruokai Yin}{yale}
\icmlauthor{Priyadarshini Panda}{yale,usc}
\end{icmlauthorlist}

\icmlaffiliation{yale}{Yale University}
\icmlaffiliation{usc}{University of Sounthern California}

\icmlcorrespondingauthor{Yuhang Li}{yuhang.li@yale.edu}

\icmlkeywords{Machine Learning, ICML}

\vskip 0.3in
]



\printAffiliationsAndNotice{\icmlEqualContribution} 

\begin{abstract}
Low-rank decomposition has emerged as an important problem in Large Language Model (LLM) fine-tuning and inference. 
Through Singular Value Decomposition (SVD), the weight matrix can be factorized into low-rank spaces optimally. 
Previously, a common practice was to decompose the weight in the activation-whitened space, and then achieve satisfying results. 
In this work, we propose Optimal-Bran Decomposition LLM (OBD-LLM), which studies the decomposition problem in the model space by utilizing second-order Hessian information.
Through a rigorous Kronecker-factorization of the Hessian, we show that the decomposition needs to consider both input and output information of the layer, and achieves much better decomposition results compared to input only method. 
Our loss-aware decomposition method involves a bi-directional whitening on the weight matrix. 
As a result, OBD-LLM is a closed-form solution for the optimal decomposition of weights in the language model. 
Remarkably, we achieve $\sim$ 20-40\% better results than previous state-of-the-art decomposition methods, the SVD-LLM. 
\end{abstract}

\section{Introduction}
\label{submission}

Large Language Models (LLMs) have become the mainstream network architecture for language generation tasks. However, the autoregressive nature of the decoder-only LLM architecture makes inference inefficient in modern hardware like, GPUs. 
This has birthed a large number of post-training compression techniques like quantization~\citep{gptq}, sparsity~\citep{sparsegpt}, and matrix decomposition~\cite{wang2024svdllm}. 

\begin{figure}[t]
\begin{center}
\centerline{\includegraphics[width=\columnwidth]{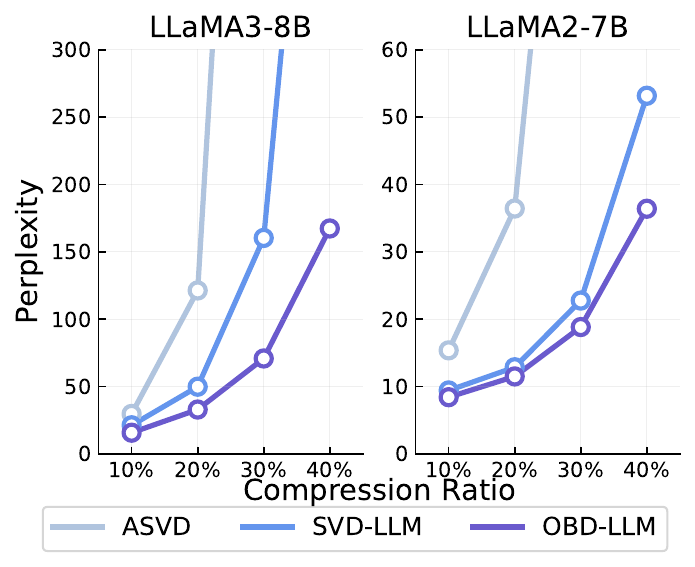}}
\vskip -0.1in
\caption{Comparison between our and previous post-training low-rank decomposition methods.}
\label{fig_intro}
\end{center}
\vskip -0.4in
\end{figure}

In this work, we focus on the matrix decomposition of weight parameters. Specifically, we are interested in the low-rank approach, where a weight matrix is typically decomposed into two low-rank matrices. Such an approximation can be made by Singular Value Decomposition (SVD) to get optimal decomposition of given ranks. 
In practice, low-rank decomposition yields many advantages. On the inference side, low-rank decomposition can directly reduce the memory and the computation costs~\citep{wang2024svdllm}. Moreover, it can also be combined with other compression techniques like pruning~\citep{zhang2025rsparse} and quantization~\citep{li2024svdquant}, where the highest singular ranks are kept in full precision and other parts are compressed.
On the training side, a low-rank adapter~\citep{hu2021lora} can significantly reduce the training memory and achieve similar performance with full fine-tuning. Decomposing the network with principled low-rank adapters may help find a better minimum of LoRA~\citep{meng2024pissa}.

More concretely, the SVD factorizes a weight matrix $\rmW\in\mathbb{R}^{m\times n}$ into two matrices $\rmB\in\mathbb{R}^{m\times r}$ and $\rmA\in\mathbb{R}^{r\times n}$, such that the weight memory and number of matrices operations is reduced by a ratio of $1-\frac{mr+nr}{mn}$. 
Hence, for a square matrix $m=n$, the kept rank must be smaller than $0.5n$ to obtain effective memory reduction.
Therefore, the singular values must contain most of the information. 

In previous literature, this information has been informed by both the weight parameters as well as the activation statistics. For example, ASVD~\citep{yuan2023asvd} uses activation scale and SVD-LLM~\citep{wang2024svdllm} uses the activation Cholesky to whiten the weight spaces. Both techniques achieve improvement over SVD. However, we argue that this type of work only focuses on input activation, i.e., the information from past layers, while ignoring the information of future layers, resulting in a suboptimal decomposition.

In this work, we propose Optimal Brain Decomposition for LLM (OBD-LLM), a principled decomposition technique that minimizes the decomposition error in the global task loss space. We start by utilizing the Hessian of the task loss function as the matrix. Then we utilize the Kronecker-factorized curvature (K-FAC) to factorize the Hessian with input activation and output gradient covariance matrix. This leads to a bi-directional whitening for weight matrix and enables more accurate decomposition.

We verify the effectiveness and efficiency of our method through direct decomposition of pretrained LLMs. For example, as shown in Fig. 1, we compare the perplexity on Wikitext2 dataset~\citep{wikitext2} of the LLaMA-2-7B~\citep{llama2} when decomposing the entire Transformer model with different methods. For both LLaMA2 and LLaMA3 models, our OBD-LLM achieves the lowest perplexity under various compression ratio on the wikitext2 dataset. We provide a more detailed experiment evaluation in \autoref{sec_exp}.

\section{Related Works}

\textbf{SVD for Language Model Compression}
Singular Value Decomposition (SVD), a widely used low-rank approximation, reduces matrix size via decomposition into two smaller low-rank matrices \citep{golub1987generalization}, making it a common choice for model compression.
For example, DRONE~\citep{chen2021drone} achieves optimal SVD compression for small language models like BERT. Yet it caches all input activations during compression, causing excessive memory usage that hinders LLM deployment. For LLMs, naive SVD on weight matrices—ignoring weight importance—induces large compression loss.
Hsu et al.~\citep{hsu2022language} proposed FWSVD, which uses Fisher information to weight parameter importance, but its complex gradient computations incur heavy resource costs for LLMs.
Another flaw of direct SVD is that activation distribution affects compression loss. Yuan et al.~\citep{yuan2023asvd} proposed ASVD, scaling weights with a diagonal matrix to normalize input channel impacts. Yet neither method links singular values directly to compression loss, so truncating small singular values may boost loss.
SVD-LLM~\citep{wang2024svdllm} pioneered loss-aware decomposition, adopting a layer-wise output reconstruction objective like GPTQ~\citep{gptq} and SparseGPT~\citep{sparsegpt}. This work advances it by leveraging the global task loss perspective.
SVD-based method can be used to other forms of compression. For example, EoRA~\citep{liu2024eora}, R-Sparse~\citep{zhang2025r} explores low-rank adapters with quantization or pruning. SVD-Quant~\citep{li2024svdquant} applied this technique on diffusion models. 
Low-Rank decomposition of the K and V projection layers can be applied to KV cache saving~\citep{changpalu}. 

\textbf{Optimal Brain Surgeon: }
 It was originally applied to
small networks with hundreds of weights~\citep{lecun1989optimal, hassibi1993optimal}. Efforts have been made to reduce the estimation complexity of Hessian on larger models, like Fisher approximation~\citep{singh2020woodfisher}, K-FAC approximation~\citep{dong2017learning}. 
Our approach utilizes the K-FAC to derive the optimal decomposition rule. 
The OBS inspired a line of post-training quantization work like \citep{adaround, brecq, gptq, li2025gptaq}. 
Recently, LLM-Surgeon~\citep{ouderaa2024the} and YAQA~\citep{tseng2025model} leverage the K-FAC to perform pruning/quantization. 

\newcommand{\rmSigma}{\mathbf{\Sigma}}
\newcommand{\rmLambda}{\mathbf{\Lambda}}
\newcommand{\rmOmega}{\mathbf{\Omega}}

\section{Preliminaries}

\begin{figure*}[t]
\centering
\begin{tikzpicture}
\node[anchor=south west,inner sep=0] (image) at (0,0) 
    {\includegraphics[width=\textwidth]{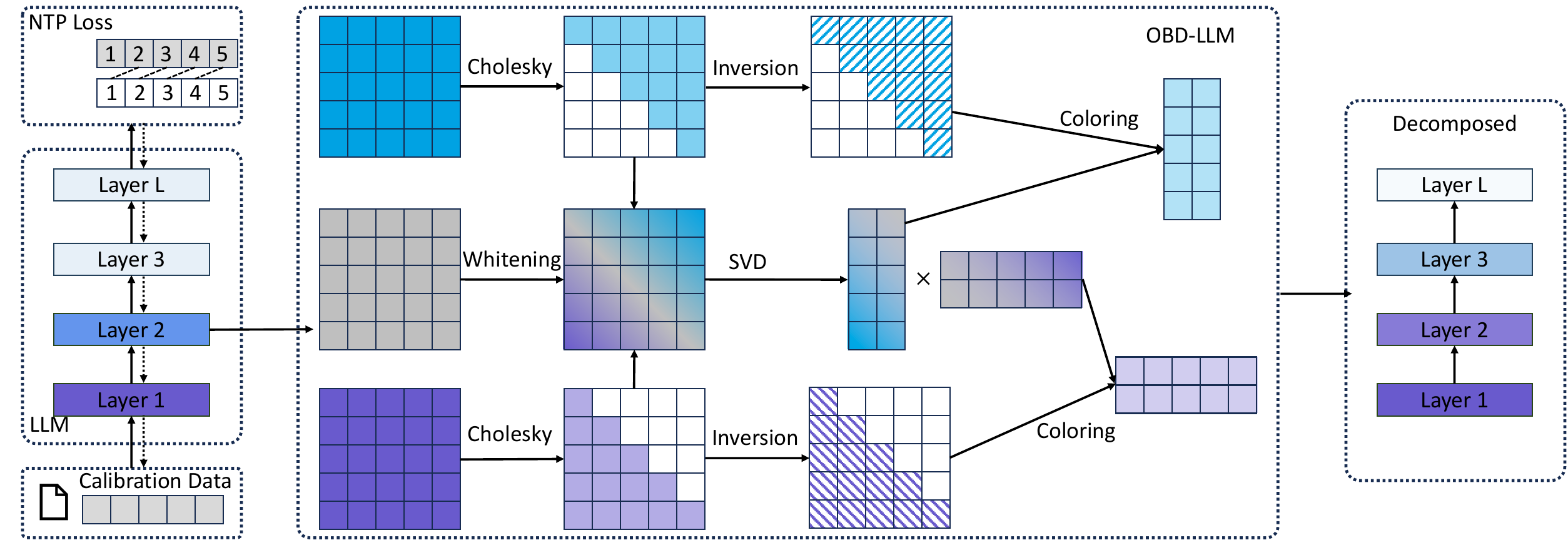}};
\begin{scope}[x={(image.south east)},y={(image.north west)}]
\node[font=\footnotesize] at (0.04,0.88) {$\mathcal{L}$};
\node[font=\footnotesize] at (0.075,0.33) {$\rmX$};
\node[font=\footnotesize] at (0.105,0.45) {$\rmG$};
\node[font=\footnotesize] at (0.32,0.05) {$\rmX\rmX^\top$};
\node[font=\footnotesize] at (0.465,0.04) {$\rmL_\rvx$};
\node[font=\footnotesize] at (0.625,0.04) {$\rmL_\rvx^{-1}$};
\node[font=\footnotesize] at (0.31,0.38) {$\rmW$};
\node[font=\footnotesize] at (0.487,0.37) {$\rmL_\rvg^\top\rmW\rmL_\rvx$};
\node[font=\footnotesize] at (0.32,0.74) {$\rmG\rmG^\top$};
\node[font=\footnotesize] at (0.465,0.73) {$\rmL_\rvg^\top$};
\node[font=\footnotesize] at (0.627,0.73) {$\rmL_\rvg^{-\top}$};
\node[font=\footnotesize] at (0.62,0.37) {$\rmU_{:,:r}\rmSigma_{:r,:r}^{1/2}$};
\node[font=\footnotesize] at (0.74,0.455) {$\rmSigma_{:r,:r}^{1/2}\rmV_{:r,:}^\top$};
\node[font=\footnotesize] at (0.79, 0.21) {$\rmA$};
\node[font=\footnotesize] at (0.79, 0.61) {$\rmB$};
\end{scope}
\end{tikzpicture}
\caption{Overview of our compression pipeline. To decompose a layer, we collect both the input activation and output activation gradient covariance matrices, and then utilize them to whiten the weight matrix for SVD. After decomposition, we utilize the inverse Cholesky to \emph{``color''} the weight back. }
\label{fig_overview}
\end{figure*}
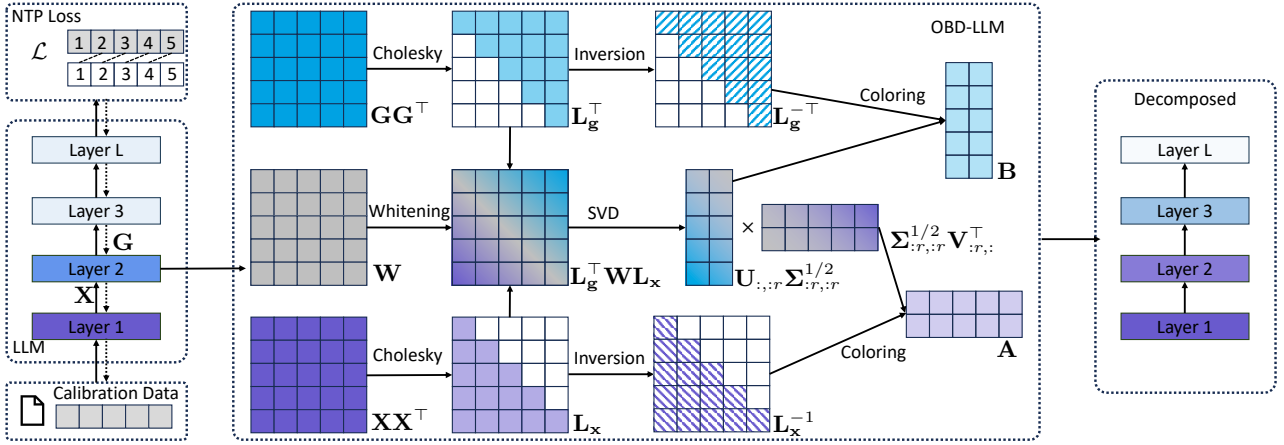



\textbf{Notations}
We adopt row-vector notation throughout this paper. Vectors and matrices are denoted by bold lowercase and uppercase letters, respectively. For instance, the linear transformation between a weight vector and input activation is expressed as:
$\rvy = \rvw \rmX$
where $\rvw \in \mathbb{R}^{1 \times n}$ denotes a row of weights (corresponding to one output channel), $\rmX \in \mathbb{R}^{n \times k}$ represents the input activation matrix, $n$ is the number of input neurons, and $k$ denotes the number of input samples.

\textbf{Singular Value Decomposition (SVD)}
Given a weight matrix $\rmW \in \mathbb{R}^{m \times n}$, SVD factorizes it into the product of three canonical matrices:
$\rmW = \rmU\rmSigma\rmV^\top$
where $\rmU = [\rvu_1, \rvu_2, \dots, \rvu_m] \in \mathbb{R}^{m \times m}$ and $\rmV = [\rvv_1, \rvv_2, \dots, \rvv_n] \in \mathbb{R}^{n \times n}$ are orthogonal matrices composed of eigenvectors of $\rmW\rmW^\top$ and $\rmW^\top\rmW$, respectively. $\rmSigma = \mathrm{diag}(\sigma_1, \sigma_2, \dots, \sigma_{\min(m,n)}) \in \mathbb{R}^{m \times n}$ is a diagonal matrix containing non-increasing singular values, which are the square roots of the eigenvalues of $\rmW\rmW^\top$ (or equivalently $\rmW^\top\rmW$). In vector form, $\rmW$ can be rewritten as a sum of rank-1 matrices:
$\rmW = \sum_{i=1}^{\min(m,n)} \sigma_i \rvu_i \rvv_i^\top$
At its core, transforming input $\rmX$ via $\rmW$ corresponds to three sequential operations: rotation via $\rmV^\top$, scaling via $\rmSigma$, and final rotation via $\rmU$.

\textbf{Low-Rank Decomposition}
To reduce memory overhead and accelerate computation of LLMs, decomposing $\rmW$ into the product of two low-rank matrices $\rmB \in \mathbb{R}^{m \times r}$ and $\rmA \in \mathbb{R}^{r \times n}$ (where $r \ll \min(m,n)$ is the decomposition rank) has become a mainstream approach. The Eckart–Young theorem~\cite{eckart1936approximation} guarantees that for $\sigma_1 > \sigma_2 > \dots > \sigma_{\min(m,n)}$, the unique solution to minimizing the Frobenius norm error $||\rmW - \hat{\rmW}||_F$ under the constraint $\mathrm{rank}(\hat{\rmW}) \le r$ is:
\begin{equation}
\hat{\rmW}^* = \rmU_{:,:r} \rmSigma_{:r,:r} \rmV^\top_{:,:r}
\end{equation}
with the error given by
\begin{equation}
||\rmW - \hat{\rmW}^*||_F^2 = \sigma_{r+1}^2 + \sigma_{r+2}^2 + \dots + \sigma_{\min(m,n)}^2
\end{equation}
Accordingly, the low-rank factors can be constructed as $\rmB = \rmU_{:,:r} \rmSigma_{:r,:r}^{1/2}$ and $\rmA = \rmSigma_{:r,:r}^{1/2} \rmV^\top_{:,:r}$. Recent work such as SVD-LLM~\citep{wang2024svdllm} extends this paradigm by incorporating activation statistics, leveraging the Cholesky factor of $\rmX\rmX^\top$ to enhance decomposition alignment with task-specific data.

\section{Optimal Brain Decomposition}

We first formulate the objective leveraging second-order information, then derive the Optimal Brain Decomposition (OBD) methodology. We further extend it to other LLM compression scenarios, e.g., quantization-aware decomposition and KV cache compression.

\subsection{Task-Loss Objective}

Unlike prior work that solely relies on input activation tensors~\citep{yuan2023asvd}, we explicitly incorporate the global task loss function into the decomposition objective. Let $\rvw = \mathrm{vec}(\rmW) \in \mathbb{R}^{1 \times mn}$ denote the row-wise flattened weight vector of $\rmW$, and $\hat{\rvw} = \mathrm{vec}(\rmB\rmA) \in \mathbb{R}^{1 \times mn}$ denote the vectorized form of the low-rank product $\rmB\rmA$ (where $\rmB \in \mathbb{R}^{m \times r}$ and $\rmA \in \mathbb{R}^{r \times n}$). We focus on minimizing the task loss change $\Delta\mathcal{L}(\rvw)$ induced by decomposition, approximated via second-order Taylor expansion:
\begin{equation}
    \Delta\mathcal{L}(\rvw) = \mathcal{L}(\rvw) - \mathcal{L}(\hat{\rvw}) \approx (\rvw - \hat{\rvw})\rmH_{\rvw}(\rvw - \hat{\rvw})^\top,
\end{equation}
where $\rmH_{\rvw}$ denotes the Hessian matrix of the loss function $\mathcal{L}$ w.r.t. $\rvw$. We retain only the quadratic term here because the gradient magnitude of the pretrained LLM is sufficiently small, leading to a negligible first-order contribution to $\Delta\mathcal{L}(\rvw)$~\citep{adaround}.

The quadratic loss expansion serves as the foundation for classic pruning paradigms: Optimal Brain Damage (OBD)~\citep{lecun1989optimal} and Optimal Brain Surgeon (OBS)~\citep{hassibi1993optimal}. From a probabilistic viewpoint, a quadratic approximation of the log-likelihood corresponds to a Gaussian approximation of the likelihood~\citep{wang2019eigendamage}—this aligns with the Laplace approximation~\citep{mackay2003information, bishop2006pattern}, where $q(\hat\rvw) = \mathcal{N}(\hat\rvw \mid \rvw + \nabla\mathcal{L}_{\rvw}, \rmH_{\rvw}^{-1})$. Here, $\rvw$ (pretrained weights) serves as the mean, and the local inverse Hessian $\rmH_{\rvw}^{-1}$ acts as the covariance matrix, capturing inter-weight correlations.

We next analyze the structural information encoded in the Hessian for LLM linear layers. For a linear layer $\rvy = \rmW\rvx$, the Hessian of the loss $\mathcal{L}$ with respect to $\rmW$ is defined as:
\begin{equation}
\begin{aligned}
\rmH_{(i,j), (h,p)} & = \frac{\partial^2\mathcal{L}}{\partial\rmW_{i,j}\partial\rmW_{h,p}} = \frac{\partial\mathcal{L}}{\partial\rmW_{i,j}}\left[ \frac{\partial\mathcal{L}}{\partial\rvy_h}\rvx_p\right] \\
& = \frac{\partial^2\mathcal{L}}{\partial\rvy_{i}\partial\rvy_{h}}\cdot \rvx_j \rvx_p 
\end{aligned}
\end{equation}
Rewriting the above in matrix form, we have
\begin{equation}
\rmH_{\rvw} = \mathbb{E} [\rvx \rvx^\top\otimes\rmH_\rvy]\in\mathbb{R}^{mn\times mn}, 
\end{equation}
where $\otimes$ denotes the Kronecker product. The expectation $\mathbb{E}$ is taken over input samples and output gradients across training/validation batches.

In practice, computing the exact full Hessian is computationally and memory-intensive, as it requires storing $O((mn)^2)$ parameters—prohibitive for large LLM layers. To address this, we approximate the Hessian with the empirical Fisher Information Matrix (FIM). For pretrained models optimized via negative log-likelihood, the Hessian is equivalent to the FIM~\citep{martens2015optimizing}, leading to:
\begin{equation}
    \rmH_\rvw = \mathbb{E}[\rvx\rvx^\top \otimes \rvg_\rvy\rvg_\rvy^\top],
    \label{eq_hessian}
\end{equation}
where $\rvg_\rvy = \frac{\partial\mathcal{L}}{\partial\rvy}$ is the first-order gradient of the layer output. 
However, it should be noted that this approximation still needs to explicitly compute the $mn\times mn$ matrix across batches, making it infeasible to operate. 

Here, we adopt the well-known K-FAC technique~\citep{martens2015optimizing} that assumes the independence of activations and derivatives, such that $ \mathbb{E}[\rvx\rvx^\top \otimes \rvg_\rvy\rvg_\rvy^\top]\approx \mathbb{E}[\rvx\rvx^\top] \otimes \mathbb{E}[\rvg_\rvy\rvg_\rvy^\top]$.
As a result, we can reconstruct the full Hessian curvature with an $m\times m$ and an $n \times n$ matrix without explicitly formulating the expensive $mn\times mn$ one.

\begin{figure}[t]
\begin{center}
\centerline{\includegraphics[width=\columnwidth]{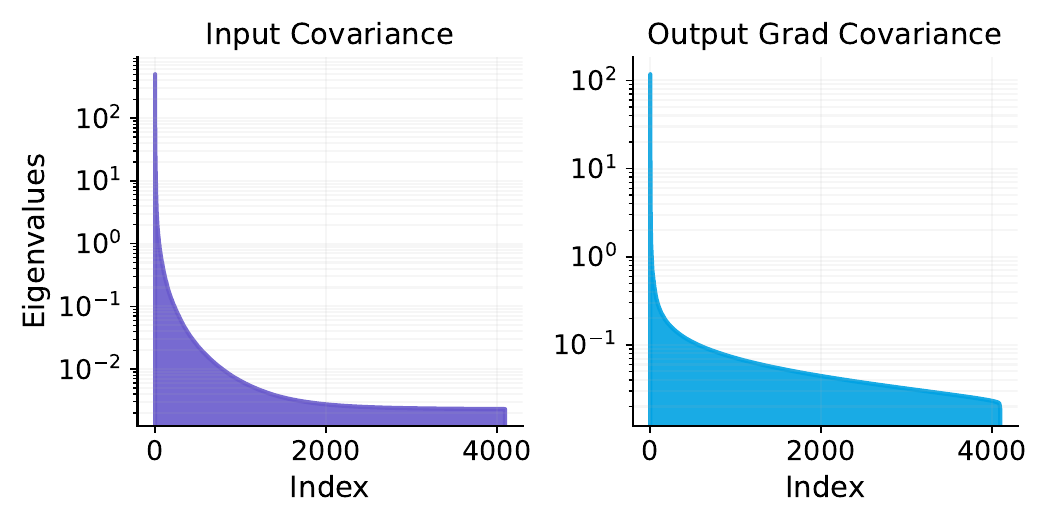}}
\vskip -0.1in
\caption{Visualization of eigenvalues of covariance matrices.}
\label{fig_cov_vis}
\end{center}
\vskip -0.4in
\end{figure}
\textbf{Comparison to Previous Work. }
A line of compression literature falls under this Hessian-based optimization. However, most of them ignore the information from $\rvg_\rvy$ and only keep the input activation $\rvx$. For example, GPTQ~\citep{gptq}, OBQ~\citep{obq} assumes that $\mathbb{E}[\rvg_\rvy\rvg_\rvy^\top]=\rmI$ and safely reduce the problem into layer-wise reconstruction objective:
\begin{equation}
\Delta\mathcal{L}(\rvw) = ||\rvw\rvx - \hat\rvw\rvx||_F^2.
\end{equation}
As a result, one only need to process the $n\times n$ Hessian matrix and thus enable fast and efficient compression algorithms. Other representative work includes SparseGPT~\citep{sparsegpt}, SVD-LLM~\citep{wang2024svdllm}. 

To demonstrate that the output gradient contains rich information as well, we visualize the eigenvalues of $\mathbb{E}[\rvx\rvx^\top] $ and $\mathbb{E}[\rvg_\rvy\rvg_\rvy^\top]$ in \autoref{fig_cov_vis}. It can be observed that both input covariance and output gradient covariance has certain eigen-dimensions that amounts for large eigenvalues, which is crucial for loss-aware decomposition.

\subsection{Optimal Solution under K-FAC}
We will now show that utilizing K-FAC can lead to optimal solution for the low-rank decomposition problem. In LLM, we use $\rmX\in\mathbb{R}^{n\times t}$ and $\rmG\in\mathbb{R}^{m\times t}$ to denote the activations and gradients with additional token dimension $t$. We define
\begin{equation}
    \rmC_\rvx = \mathbb{E}\left[\rmX\rmX^\top\right],\ \  \rmC_\rvg = \mathbb{E}\left[\rmG\rmG^\top\right],
\end{equation}
as the covariance matrices of $\rmX$ and $\rmG$\footnote{We call it the covariance matrix by assuming zero mean, though it might not represent the actual ``covariance". }. 
Hence, we can rewrite the Hessian as $\rmH_{\rvw} = \rmC_\rvx \otimes \rmC_\rvg$ with the K-FAC approximation. 
Now, denoting the weight error as $\Delta\rmW = \rmW-\hat{\rmW}$, we can rewrite the loss objective $\Delta\mathcal{L}(\rvw)$ as 
\begin{equation}
\begin{aligned}
    \Delta\mathcal{L}(\rvw) &  = 
    \mathrm{vec}(\Delta\rmW)\rmH_\rvw \mathrm{vec}({\Delta\rmW})^\top \\
    &  = \mathrm{vec}(\Delta\rmW)(\rmC_\rvx\otimes\rmC_\rvg) \mathrm{vec}(\Delta\rmW)^\top \\
    &  = \mathrm{vec}(\Delta\rmW) \mathrm{vec}(\rmC_\rvg\Delta\rmW\rmC_\rvx)^\top \\
    & = \mathrm{tr}(\Delta\rmW^\top\rmC_\rvg\Delta\rmW\rmC_\rvx),
\end{aligned}
\end{equation}
using the Kronecker product's property $(\rmA\otimes\rmB)\mathrm{vec}(\rmC) = \mathrm{vec}(\rmB\rmC\rmA^\top)$ and the fact that covariance matrices are symmetric. 

To solve this problem, we apply the Cholesky Decomposition to both covariances, given by
\begin{equation}
    \rmC_\rvx = \rmL_\rvx\rmL_\rvx^\top, \ \ \ \rmC_\rvg = \rmL_\rvg\rmL_\rvg^\top,
\end{equation}
where $\rmL_\rvx\in\mathbb{R}^{n\times n},\rmL_\rvg\in\mathbb{R}^{m\times m} $ are the lower-triangular Cholesky factor. To this end, we can rewrite the loss objective as
\begin{subequations}
\begin{align}
\Delta\mathcal{L}(\rmW) & = \mathrm{tr}(\Delta\rmW^\top\rmC_\rvg\Delta\rmW\rmC_\rvx) \\
& = \mathrm{tr}(\Delta\rmW^\top\rmL_\rvg\rmL_\rvg^{\top}\Delta\rmW\rmL_\rvx\rmL_\rvx^{\top}) \\
& = \mathrm{tr}(\rmL_\rvg^{\top}\Delta\rmW \rmL_\rvx\rmL_\rvx^{\top}\Delta\rmW^\top\rmL_\rvg) \label{eq_tr_cycle}\\
& = ||\rmL_\rvg^{\top}\Delta\rmW \rmL_\rvx||_F^2, \label{eq_tr_norm}
\end{align}
\end{subequations}
where \autoref{eq_tr_cycle} uses the cyclic property of the trace, and \autoref{eq_tr_norm} uses the property that $\mathrm{tr}(\rmA\rmA^\top) = ||\rmA||_F^2$.

To this end, we can construct a weight matrix $\tilde{\rmW} = \rmL_\rvg^{\top}\rmW \rmL_\rvx$ in the whitened space, and then our objective is to find the optimal low-rank approximation of the whitened weight. This can be solved by applying SVD, given by:
\begin{equation}
\tilde{\rmW} = \rmU\rmSigma\rmV^\top.
\end{equation}
To obtain the low rank matrices in the original space, we need the inverse transformation to both sides:
\begin{equation}
\rmB = \rmL_\rvg^{-\top}\rmU_{:,:r}\rmSigma_{:r,:r}^{1/2}, \ \ \ \rmA = \rmSigma_{:r,:r}^{1/2}\rmV^\top_{:r,:}\rmL_\rvx^{-1}.
\end{equation}
Essentially, we show that the Kronecker-factorized Hessian matrix indicates two whitening spaces, the input activation $\rmX$ and the output derivatives $\rmG$. By enforcing whiteing operations into both spaces, e.g.,
$(\rmL_\rvx^{-1}\rmX)(\rmL_\rvx^{-1}\rmX)^\top = \rmL_\rvx^{-1}\rmX\rmX^\top\rmL_\rvx^{-\top} = \rmI$, we make the input activation and output derivatives orthogonal. Notably, $\rmL_\rvx$ orthogonalizes the input channel dimension while, $\rmL_\rvg$ orthogonalizes the output channel dimension. This bi-directional whitening matrix is the key to OBD-LLM.
The operations are shown in \autoref{fig_overview}.

\textbf{Independence Assumption in K-FAC. }
One may ask how much information will be lost in the K-FAC approximation of the Hessian. Although \cite{martens2015optimizing} states that most information of the Hessian is preserved in the Kronecker structure, while the correlation of $\rmX$ and $\rmG$ is minimal. Here, we provide a empirical justification that K-FAC is pretty accurate. 

\begin{figure}[t]
\begin{center}
\centerline{\includegraphics[width=\columnwidth]{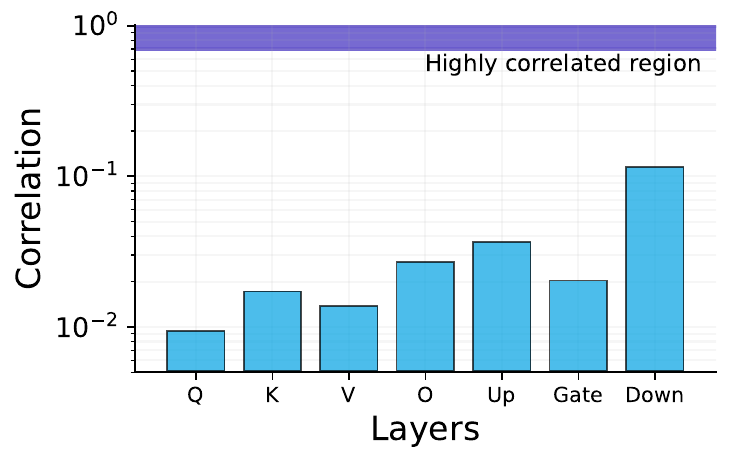}}
\vskip -0.2in
\caption{Visualization of correlation factor across all projection layers in LLaMA-3-8B.}
\label{fig_corr_hist}
\end{center}
\vskip -0.4in
\end{figure}

We define a correclation factor $\rho$ by
\begin{equation}
\rho = \frac{||\rmX\rmG^\top||_F^2}{\sqrt{||\rmX\rmX^\top||_F^2}\sqrt{||\rmG\rmG^\top||_F^2}}\in[0, 1].
\end{equation}
As this correlation factor approaches 1, the input activation and output gradient activation are becoming more dependent and K-FAC approaximation would fail to estimate the actual Hessian information. 
We provide empirical results of every projection layer in LLaMA-3-8B. As demonstrated in \autoref{fig_corr_hist}, all layers have $\le 0.1$ correlation between $\rmX$ and $\rmG$, which indicates the practical espressiveness of K-FAC.

\subsection{Extension to Other Forms of Compression}

We also show that our framework can be generalized to two additional compression scenarios: (1) Sparse/Quantized + Low Rank Adapters, and (2) Low-Rank KV-Cache. 

\textbf{Sparse/Quantized + Low Rank Adapters. }
In this case, we define the $\Delta\rvw = \rvw - \hat\rvw - \mathrm{vec}(\rmB\rmA)$ where $\hat\rvw$ is the quantized or sparse weights after applying GPTQ or other kinds of compression algorithms.  
\cite{liu2024eora} have shown that tiny low-rank adapters with 128 rank can greatly compensate the performance drop after pruning or quantization. For this setup, we transform the weight compression error $\tilde{\rmW} = \rmL_\rvg^\top (\rmW-\tilde{\rmW})\rmL_\rvx$, and apply SVD to obtain the OBD-LLM adapters.

\textbf{Low-Rank KV Cache. }
Inspired by Palu~\citep{changpalu}, the decomposition of K and V projection layers into low-rank matrices can effectively reduce the KV Cache memory by storing $\rmA\rmX$ as cache and reconstructing it by $\rmB$. 
For V projection layer, since the KV cache and attention is parallelly processed in each head, a unified $\rmC_\rvg$ would introduce intra-head information for compression, and thus is not compatible with our current strategy. To solve this, we need to parallelly process each head's unique $\rmC_\rvg$. More concretely, for each head's weights $\rmW\in\mathbb{R}^{h\times n}$, we use the shared $\rmC_\rvx$ and the unique $\rmC_\rvg$ of that head to decompose the weights. 

For low-rank K cache, the challenge is that decomposing the K projection layer will need to reconstruct the RoPE embedding at each decoding step. Although \citep{changpalu} provide a decoding kernel for RoPE, other lines of work chose to directly decompose the K cache post-RoPE using PCA. Our OBD-LLM framework can be extended to this PCA framework. Concretely, denoting the K cache as $\rmK = \mathrm{RoPE}(\rmW_\rmK \rmX)$, post-RoPE low-rank compression seeks to use PCA that computes the eigendecomposition of $\rmK\rmK^\top = \rmU\rmLambda\rmU^\top$. As such, we can apply $\rmU_{:, :r}$ to map the K cache into the top-$r$ principle components and perserve the variance as much as possible. 

For our OBD-LLM, we adopt Hessian-based compression where we are interested in 
\begin{equation}
\min_{\Delta\rmK} \mathcal{L}(\Delta\rmK) = \min_{\Delta\rmK} \mathrm{vec}(\Delta\rmK)\rmH_{\rmK} \mathrm{vec}(\Delta\rmK)^\top,
\end{equation}
where we use the empirical Fisher to approximate Hessian $\rmH_{\rmK} = \mathbb{E}[\rmG_\rmK\rmG_\rmK^\top]$. 
Similarly, by factorizing the Hessian with Cholesky decomposition $\rmH_\rmK=\rmL_\rmK \rmL_\rmK^\top$, we can whiten the K cache by $\rmL_\rmK$ and then apply the PCA. 
Formally, denoting $\tilde{\rmK} = \rmK\rmL_\rmK$, we apply the PCA $\tilde{\rmK}\tilde{\rmK}^\top = \rmU\rmLambda\rmU$ to select the top-r components. To reconstruct the KV cache, we will utilize the inverse of Cholesky  $\rmLambda^{1/2}_{:r, :}\rmU\rmL_\rmK^{-1}$ as well.

\newcommand{\cmark}{\ding{51}}
\newcommand{\xmark}{\ding{55}}

\begin{table*}[t]
\setlength{\tabcolsep}{5pt}
\small
\centering
\caption{{Low-rank decomposition results of LLaMA2/3 Models}. We use 128 samples from C4 dataset to compress the pretrained LLM by 10\%$\sim$40\% and report the wikitext2 and C4 perplexity. }
\begin{adjustbox}{max width=\linewidth}
\begin{tabular}{llcccccccccccccc}
\toprule
\multirow{2}{*}{\textbf{Model}} & \multirow{2}{*}{\textbf{Method}}  & \multicolumn{2}{c}{\textbf{10\%}} & \multicolumn{2}{c}{\textbf{20\%}} & \multicolumn{2}{c}{\textbf{30\%}} & \multicolumn{2}{c}{\textbf{40\%}}\\
\cmidrule{3-4} \cmidrule{5-6} \cmidrule{7-8} \cmidrule{9-10}
 & & \textbf{Wiki2} & \textbf{C4}  & \textbf{Wiki2} & \textbf{C4}  & \textbf{Wiki2} & \textbf{C4}   & \textbf{Wiki2} & \textbf{C4} \\
\midrule
& SVD & nan & nan & nan & nan & nan & nan & nan & nan \\
& FWSVD & 3.0e5 & 3.1e5  & 3.7e5 & 3.7e5 & 4.4e5 & 4.7e5 & 6.1e5 & 7.1e5 \\
LLaMA3-8B& ASVD~\citep{yuan2023asvd}  & 29.68 & 37.62 &  121.3 & 127.2 & 927.2 & 622.4 & 2253 & 1381 \\
& SVD-LLM~\citep{wang2024svdllm}  & 21.23 & 18.49 & 49.88 & 31.25 & 160.3 & 73.14 & 679.8 & 238.3 \\
\rowcolor{gray!20}\cellcolor{white} & OBD-LLM (Ours)  & \bfseries 15.73 & \bfseries 16.42 & \bfseries 32.92 & \bfseries 24.72 & \bfseries 70.83 & \bfseries 44.36 & \bfseries 167.3 & \bfseries 94.46 \\
\midrule
& SVD & 1.6e5 & nan & 1.8e5 & nan &  3.1e5 &  nan & nan & nan \\
& FWSVD & 1.6e4 & 2.0e4 & 1.8e4 & 2.6e4 & 3.0e4 & 3.8e4 & 3.9e4 & 5.6e4 \\
LLaMA2-7B  & ASVD~\citep{yuan2023asvd} & 15.37 & 17.90 & 36.42 & 36.74 & 134.9 & 119.86 & 705.3 & 608.2\\
& SVD-LLM~\citep{wang2024svdllm} & 9.42 & 11.16 & 12.86 & 13.72 & 22.76 & 19.78 & 53.16 & 34.34 \\
\rowcolor{gray!20}\cellcolor{white} & OBD-LLM (Ours) & \bfseries 8.42 & \bfseries 10.14 & \bfseries 11.50 & \bfseries 12.14 & \bfseries 18.83 & \bfseries 16.61 & \bfseries 36.39 & \bfseries 27.37 \\
\bottomrule
\end{tabular}
\end{adjustbox}
\label{tab_llama}
\end{table*}

\begin{table*}[t]
\setlength{\tabcolsep}{5pt}
\small
\centering
\caption{{Low-rank decomposition of LLaMA2/3 Models (20\% compression ratio)}. We report 8 different zero-shot downstream task accuracy and the average performance. }
\begin{adjustbox}{max width=\linewidth}
\begin{tabular}{llcccccccccccccc}
\toprule
\textbf{Model} & \textbf{Method} & \textbf{PiQA} & \textbf{Arc E}  & \textbf{Arc C} & \textbf{HellaSwag}  & \textbf{Winogrande} & \textbf{BoolQ}   & \textbf{OBQA} & \textbf{SiQA} & \textbf{Avg} \\
\midrule
& Full Rank & 80.63 & 77.74 & 53.50 & 79.12 & 73.32 & 81.19 & 44.80 & 33.01 & 65.41 \\
& FWSVD & 51.58 & 25.21 & 27.73 & 26.22 & 50.12 & 47.40 & 29.20 & 33.57 & 36.38 \\
LLaMA3-8B & ASVD & 63.38 & 43.86 & 27.65 & 41.95 & 56.75 & 66.45 & 32.40 & 33.42 & 45.73 \\
& SVD-LLM  & 72.25 & 54.84 & 32.68 & 57.51 & 66.06 & 68.29 & 37.00 & 32.91 & 52.69 \\
\rowcolor{gray!20}\cellcolor{white} & OBD-LLM (Ours)  & \bfseries 73.72 & \bfseries 59.47 & \bfseries 34.90 & \bfseries 59.47 & \bfseries 67.96 & \bfseries 68.56 & \bfseries 36.80 & \bfseries 32.96 & \bfseries 54.23 \\
\midrule
& Full Rank & 79.05  & 74.54 & 46.25 & 75.99 & 68.90 & 77.71 & 44.20 & 32.91 & 62.44  \\
& FWSVD & 51.58 & 25.21 & 27.73 & 26.22 & 50.12 & 47.40 & 29.20 & 33.57 & 36.38 \\
LLaMA2-7B & ASVD & 63.82 & 47.22 & 29.61 & 46.6  & 57.22 & 59.60 & 30.40 & 31.99 & 45.81 \\
& SVD-LLM  & 72.63 & 52.61 & 32.25 & 58.68 & 64.17 & 64.31 & 36.80 & 33.62 & 51.88 \\
\rowcolor{gray!20}\cellcolor{white} & OBD-LLM (Ours)  & \bfseries 74.37 & \bfseries 58.67 & \bfseries 34.90 & \bfseries 61.34 & \bfseries 66.22 & \bfseries 70.03 & \bfseries 36.20 & \bfseries 34.14 & \bfseries 54.48 \\
\bottomrule
\end{tabular}
\end{adjustbox}
\label{tab_llama_acc}
\end{table*}

\subsection{Implementation}

Similar to previous literature \citep{wang2024svdllm}, our OBD-LLM only requires a small number of input sequence to collect the neccessary information. In practice, we select 128 input sequence to form a calibration dataset. For input activation covariance, we directly collect the activation of each layer. For output gradient covariance, we compute the loss function with next-token prediction loss, given by
\begin{equation}
\mathcal{L}(\theta, \rvx) = -\sum_{t=1}^{T-1} \rvx_{t+1} \mathrm{Softmax}(\log (\theta(\rvx_t))),
\end{equation}
where $\theta$ denotes the model parameters and $T$ is the sequence length. In practice, we can further add a temperature to the logit of network output, see \autoref{sec_ablation} for discussion on this. 

For decomposition, we adopt the standard PyTorch function to execute the SVD function. Furthermore, given that we need to transform the whitened matrix back to original space, the inverse of Cholesky needs to be calculated. Here we use the ``\texttt{triangular solve}" function that computes 
\begin{equation}
\rmL_\rvg^\top \rmB = \rmU_{:,:r}\rmSigma^{1/2}_{:r,:r},\ \  \rmA\rmL_{\rvx} = \rmSigma^{1/2}_{:r,:r}\rmV^\top_{:r, :},
\end{equation}
which can be solved in $O(n^2)$ time and avoids computing the inverse of the Cholesky explicitly.

\section{Experiments}
\label{sec_exp}

\subsection{Setup}
We implement OBD-LLM using Hugging Face~\citep{wolf2019huggingface} on top of the PyTorch framework~\citep{paszke2019pytorch}. 
We select 128 2048-token samples from c4 dataset~\citep{c4} to compute the input covariance and output gradient covariance. 
For better numerical stability and better-conditioned covariance, we add 10\% average diagonal value dampening to each covariance matrix. 
We begin our program by first saving all the covariance matrices. This process takes roughly 3 minutes on the model we tested, which is relatively small compared to decomposition runtime.

We first compare the standard low-rank decomposition performance by directly truncating the lowest ranks. Then, we compare other forms of compression, e.g., sparse or quantized + low rank adapters and low-rank KV cache. 

\subsection{Comparison of Low-Rank Decomposition}
We verify our experiments on large language transformer architectures, including LLaMA2-7B~\citep{llama2} and LLaMA3-8B~\citep{llama31}.
Here, we perform low-rank decomposition in fixed compression ratios, ranging from 10\% to 40\%. 
Note that for simplicity and fair comparison, we apply uniform rank truncation ratio to all layers. 
We compare several existing approaches including standard SVD, FWSVD~\citep{hsu2022language}, ASVD~\citep{yuan2023asvd}, SVD-LLM~\citep{wang2024svdllm}.

\begin{table*}[t]
\setlength{\tabcolsep}{5pt}
\small
\centering
\caption{Comparison of quantization/pruning + low-rank adapter compensation. We compare No LRC (no low-rank compensation), SVD compensation, EoRA and our OBD-LLM. The adapter has 128 rank. }
\begin{adjustbox}{max width=\linewidth}
\begin{tabular}{lclccccccccccccc}
\toprule
\multirow{2}{*}{\textbf{Model}} & \multirow{2}{*}{\textbf{Rank}}& \multirow{2}{*}{\textbf{Method}} & \multicolumn{3}{c}{\textbf{W3 Quant}} & \multicolumn{3}{c}{\textbf{2:4 Sparsity}} \\
\cmidrule{4-6} \cmidrule{7-9}
 & & & \textbf{Wiki2} & \textbf{C4} & \textbf{Avg Acc} &  \textbf{Wiki2} & \textbf{C4} & \textbf{Avg Acc}   \\
\midrule
& 0 & No LRC & 18.12 & 24.83 & 49.84 & 16.32 & 23.06 & 52.73 \\
\cmidrule{2-9}
& 64 & SVD & 12.87 & 18.50 & 51.87 & \bfseries 15.22 & 21.61 & 53.68\\
& 64 & EoRA \citep{liu2024eora} & 12.80 & 18.57 & 51.52 & 15.74 & 21.89 & 53.73\\
\rowcolor{gray!20}\cellcolor{white}LLaMA3-8B & 64 & OBD-LLM & \bfseries 12.36 & \bfseries17.89 & \bfseries53.14 & 15.24 &\bfseries 21.52 &\bfseries 53.82 \\
\cmidrule{2-9}
& 128 & SVD & 12.06 & 17.76 & 52.97 & 14.53 & 20.63 & 54.54\\
& 128 & EoRA \citep{liu2024eora} & 11.42 & 17.45 & 53.10 & 14.36 & 20.52 & 55.14 \\
\rowcolor{gray!20}\cellcolor{white}& 128 & OBD-LLM & \bfseries11.42 &\bfseries 16.87 &\bfseries 54.41 &\bfseries 14.27 & \bfseries20.32 & \bfseries55.50\\
\midrule
& 0 & No LRC & 8.37 & 10.80 & 54.59 & 10.92 & 13.85 &  53.75 \\
\cmidrule{2-9}
\multirow{2}{*}{LLaMA2-7B} & 64 & SVD & 7.52 & 9.94 & 56.46 & 10.28 & 13.12 & 54.28 \\ 
& 64 & EoRA \citep{liu2024eora} & 7.59 & 10.03 & 56.22 & 10.34 & 13.08 & 54.25\\
\rowcolor{gray!20}\cellcolor{white}& 64 & OBD-LLM & \bfseries7.39 &\bfseries 9.82 & \bfseries57.30 &\bfseries 10.20 & \bfseries13.01 &\bfseries 54.41 \\
\bottomrule
\end{tabular}
\vspace{-1em}
\end{adjustbox}
\label{tab_svd_quant}
\end{table*}

\textbf{Perplexity Evaluation. }We first compare the perplexity performance in \autoref{tab_llama}. 
We note that on both LLaMA2 and LLaMA3 models, SVD-LLM, which leverages the input activation Cholesky, is the current state-of-the-art. Our OBD-LLM further improves the perplexity performance. 
On LLaMA2-7B with 20\% compression ratio, our method improves the perplexity from 12.86 to 11.50 on the wikitext2 dataset.
For LLaMA3 models, low-rank decomposition becomes more challenging, a common observation in other compression schemes as well~\citep{li2025gptaq}. For example, SVD-LLM has nearly 50 perplexity on wikitext2 dataset, much higher than the LLaMA2 model. 
Our method can improve it to 32.9. It is also worthwhile to note that other methods like ASVD and FWSVD fail to capture the model curvature for LLaMA3, resulting in much higher perplexity, e.g., 121 for ASVD and crashed performance for FWSVD. 
We also test more extreme compression rates like 40\%. Under this ratio, all existing method will lead to a huge gap against full-rank methods. And our OBD-LLM demonstrates ability to brigde the gap between decomposed model and full-rank model. 

\textbf{Downstream Accuracy Evaluation. }
We show the performance on six downstream tasks: PiQA~\citep{piqa}, ARC easy / challenged~\citep{arc}, Hellaswag~\cite{hellaswag}, Winogrande~\cite{sakaguchi2021winogrande}, BoolQ~\cite{boolq}, OBQA~\citep{OpenBookQA2018}, and SiQA~\citep{sap2019social}.
We test various methods under the 20\% compression ratio and provide the full-rank baseline performance. 
The results are shown in \autoref{tab_llama_acc}. 
With SVD-LLM, the decomposed model has 12.7\% and 10.6\% gap with the full rank 8B and 7B models, respectively. Our OBD-LLM can effectively reduce these gaps to 11.2\% and 8.0\%.  
Again, other methods like ASVD and FWSVD incur a much more significant drop on the downstream task accuracy. 

\vspace{-1em}
\subsection{Comparison of Low-Rank Compensation}

Next, we verify the performance of using low-rank adapters in a quantized or sparse model. We select the two most common compression algorithms: GPTQ~\citep{gptq} for quantization and SparseGPT~\citep{sparsegpt} for pruning. 
We apply 3-bit per-channel weight quantization and 2:4 pruning on the LLaMA2-7B model and LLaMA3-8B model. 
For comparison, we use (1) standard SVD on the residual of the weights, i.e. $(\rmW-\tilde{\rmW})$, (2) EoRA~\citep{liu2024eora} which whitens the residual weight with the input activation covariance. 
Similarly, our method considers bi-directional whitening. 
For low-rank compensation, we have to make it as lightweight as possible so that it only incurs negligible computation. Therefore, we add either 64 or 128 ranks to the compensate the compression. 

The results are demonstrated in \autoref{tab_svd_quant}, where we provide both the perplexity and average downstream task accuracy. 
It can be observed that ``No LRC", which stands for No Low-Rank Compensation, will significantly degrade the performance of full-precision due to aggressive quantization/pruning. Unlike No low-rank compression, the standard SVD can achieve a decent performance since the majority of compressed weight is obtained with input covariance $\rmC_{\rvx}$. 
We also find that EoRA does not bring a significant improvement over SVD baseline. Again, we attribute the reason to GPTQ/SparseGPT as they already optimize the weight with $\rmC_{\rvx}$. Our method, on the contrary, brings additional information to the compressed model. Therefore, we can find that OBD-LLM can significantly improve the performance. For example, OBD-LLM achieves 57.30 average accuracy with 1.1\% improvement over SVD-LLM using only 64 ranks.

\begin{table}[t]
\setlength{\tabcolsep}{5pt}
\small
\centering
\caption{Comparison of KV Cache Compression. }
\begin{adjustbox}{max width=\linewidth}
\begin{tabular}{l clcc }
\hline\hline
 \textbf{Model} & \textbf{Comp. Ratio} & \textbf{Method} & \textbf{LongBench} \\
\hline
  & 0\% & Full-Rank  & 36.32 \\
  \cline{2-4}
 & \multirow{2}{*}{30\%} & PaLU  & 35.45 \\
 LongChat-7B-v1.5 & & OBD-LLM & \bfseries 35.79 \\
 \cline{2-4}
  & \multirow{2}{*}{50\%} & PaLU  & 30.82 \\
 & & OBD-LLM & \bfseries 32.27 \\
\hline\hline
\end{tabular}
\vspace{-2em}
\end{adjustbox}
\label{tab_kvcache}
\end{table}

\subsection{Comparison of Low-Rank KV Cache}

Now, we compare the performance of OBD-LLM with KV cache compression. 
We choose baseline as PaLU~\citep{changpalu}. We evaluate on the LongChat-7B model with LongBench~\citep{longbench} up to 32k context length. We compress the KV cache by 30\% or 50\%. The rest of the setup follows the PaLU experiments. 
As demonstrated in \autoref{tab_kvcache}, our OBD-LLM improves PaLU significantly, espcially on the 50\% compression ratio. The LongBench average accuracy is increased from 30.8 to 32.3. 

\subsection{Ablation Study }

\label{sec_ablation}

In this section, we conduct ablation study of OBD-LLM compression algorithm. 
The proposed algorithm comprises two Cholesky factor for whitening, $\rmL_\rvx$ and $\rmL_\rvg^\top$, which can be viewed as information from past layer and information from future layer, respectively. 
Therefore, we test the performance of applying these two terms \emph{individually} and observe how they contribute to the final performance. 
We conduct experiments on 20\% compression ratio with LLaMA3-8B. The results are demonstrated in \autoref{tab_ablation}. 
Note that if we do not empower any update to weights, the method will reduce to SVD; and SVD-LLM is the case where we only apply $\rmL_\rvx$.  
Interestingly, we can find that solely applying the first term or second term can increase the decomposition performance a lot compared to SVD. 
While applying the first term obtains a better perplexity score, the average accuracy when applying the second term individually is much better, resulting in 2\% higher performance. 
Combining both terms, which is our OBD-LLM, the decomposition performance can be further improved. This result suggests that information from the future should be taken into account during decomposition.

\begin{table}[t]
\setlength{\tabcolsep}{5pt}
\small
\centering
\caption{{Ablation study of decomposition objectives}. }
\begin{adjustbox}{max width=\linewidth}
\begin{tabular}{l llcc }
\hline\hline
 \textbf{Method} & \textbf{Objective} & \textbf{Wiki2$(\downarrow)$} & \textbf{C4$(\downarrow)$} & \textbf{Avg$(\uparrow)$} \\
\hline
 SVD & $||\Delta\rmW||_F^2$ & 1.8e5 & nan & 35.59 \\
 SVD-LLM & $||\Delta\rmW\rmL_{\rvx}||_F^2$  & 49.88 & 31.25 & 52.69 \\
 OBD-LLM$'$ & $||\rmL_\rvg^{\top}\Delta\rmW||_F^2$  & 101.7 &  88.83 & 39.56 \\
\rowcolor{gray!20}OBD-LLM & $||\rmL_\rvg^{\top}\Delta\rmW\rmL_\rvx||_F^2$ & \bfseries 32.92 &  \bfseries 24.72 & \bfseries 54.23 \\
\hline\hline
\end{tabular}
\end{adjustbox}
\vspace{-1em}
\label{tab_ablation}
\end{table}

\begin{table}[t]
\setlength{\tabcolsep}{5pt}
\small
\centering
\caption{{Ablation study of temperature for loss function}. }
\begin{adjustbox}{max width=\linewidth}
\begin{tabular}{l cccc }
\hline\hline
 \textbf{Method} & \textbf{Temperature} & \textbf{Wiki2$(\downarrow)$} & \textbf{C4$(\downarrow)$} & \textbf{Avg$(\uparrow)$} \\
\hline
 & 0.5 & \bfseries 11.05 & 12.21 & \bfseries 54.48 \\
 OBD-LLM & 1.0  & 11.50 & \bfseries 12.14 & 54.23 \\
 & 2.0  & 11.15 & 12.45 & 53.91 \\
\hline\hline
\end{tabular}
\end{adjustbox}
\vspace{-1em}
\label{tab_ablation_temp}
\end{table}

Next, we ablate the loss function, which will affect the gradient covariance estimation. We can apply a temperature on the logits of the LLM output for calculating the loss function. A large temperature will flatten the probability across the vocabulary, which will have more uniform distribution and vice versa. We perform $T=\{0.5, 1, 2.0\}$ on LLaMA3-8B, and summarize the results in \autoref{tab_ablation_temp}. It can be observed that a $T=0.5$ obtains best downstream task accuracy and wikitext2 perplexity, indicating that a sharper loss surface will contribute to a better estimation of loss curvature.




\begin{figure}[t]
\begin{center}
\centerline{\includegraphics[width=\columnwidth]{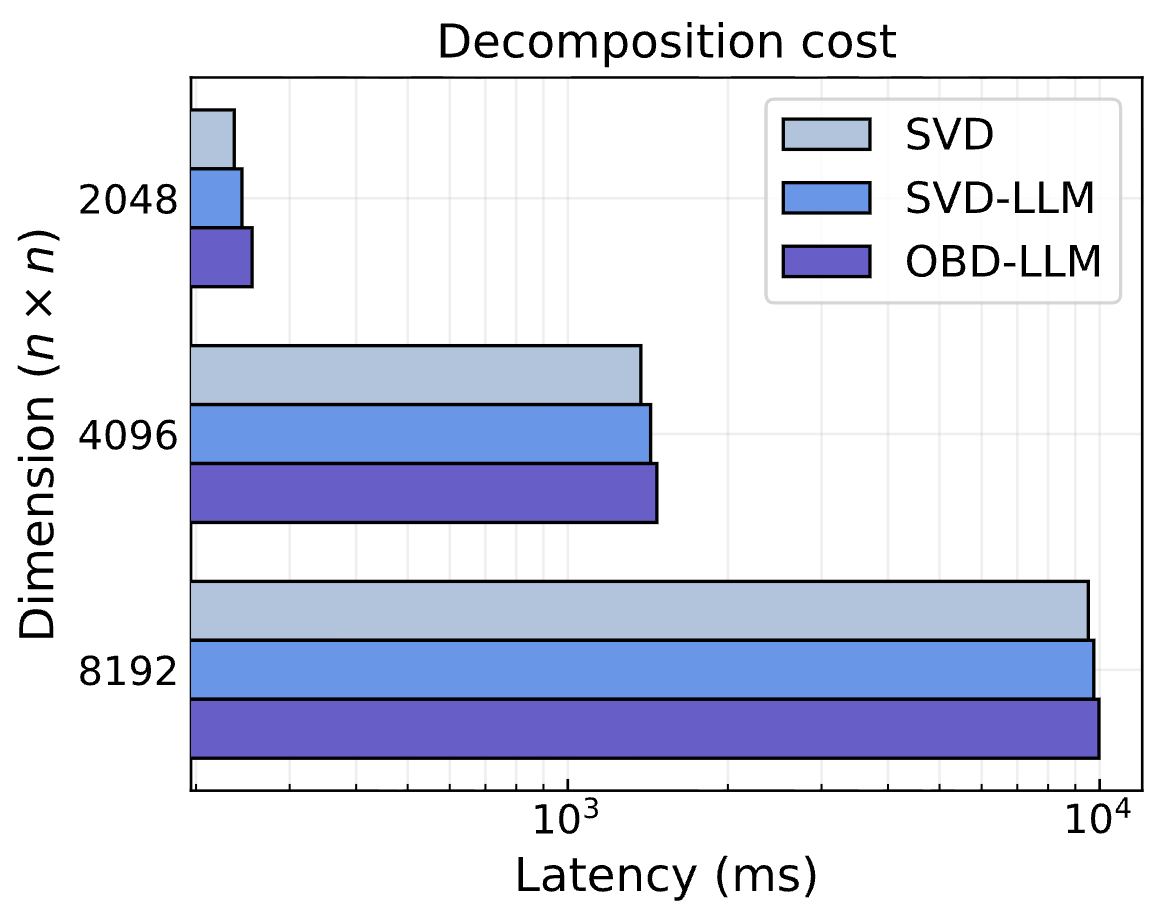}}
\vskip -0.1in
\caption{Latency comparison of decomposition runtime.}
\label{fig_decomp_time}
\end{center}
\vskip -0.3in
\end{figure}

\subsection{Efficiency Analysis}

We also compare the algorithm runtime of different decomposition methods. We select the standard SVD, SVD-LLM and our OBD-LLM and decompose a $n\times n$ matrix where $n=\{2048, 4096, 8192\}$. 
We measure the latency on one A100 and average the run by 10 repeated runs. 
Theoretically, SVD-LLM and OBD-LLM will add complexity to the decomposition process. 
However, as shown in \autoref{fig_decomp_time}, both SVD-LLM and our OBD-LLM adds negligible (4\%$\sim$7\%) latency to the overall decomposition process.
This is due to the well-optimized Cholesky and triangular solve kernel on GPUs. The overall latency is dominated by the SVD process, which requires $O(n^3)$ iteration to find the solution.

\section{Conclusion}

This paper proposes Optimal Brain Decomposition LLM, a loss-aware low-rank decomposition method for LLMs. By integrating global task loss into the objective and simplifying Hessian computation via Kronecker factorization, OBD addresses the limitations of existing SVD-based methods. It establishes an explicit connection between decomposition operations and compression loss, achieving a better balance between model size reduction and performance preservation compared to FWSVD, ASVD, and SVD-LLM. OBD’s layer-wise adaptability also extends to multi-scenario compression like quantization. Future work will explore its generalization to multi-modal LLMs and integration with dynamic low-rank adjustment to further enhance efficiency in real-world deployment.

\vspace{-1em}
\section*{Impact Statement}

This paper presents work whose goal is to advance the field of 
Machine Learning. There are many potential societal consequences 
of our work, none which we feel must be specifically highlighted here.

\nocite{langley00}

\bibliography{example_paper}
\bibliographystyle{icml2026}



\end{document}

%% file: math_commands.tex
\usepackage{amsmath,amsfonts,bm}

\def\eqref#1{equation~\ref{#1}}

\def\1{\bm{1}}

\def\rvg{{\mathbf{g}}}

\def\rvu{{\mathbf{i}}}

\def\rvu{{\mathbf{u}}}
\def\rvv{{\mathbf{v}}}
\def\rvw{{\mathbf{w}}}
\def\rvx{{\mathbf{x}}}
\def\rvy{{\mathbf{y}}}

\def\rmA{{\mathbf{A}}}
\def\rmB{{\mathbf{B}}}
\def\rmC{{\mathbf{C}}}

\def\rmG{{\mathbf{G}}}
\def\rmH{{\mathbf{H}}}
\def\rmI{{\mathbf{I}}}

\def\rmK{{\mathbf{K}}}
\def\rmL{{\mathbf{L}}}

\def\rmU{{\mathbf{U}}}
\def\rmV{{\mathbf{V}}}
\def\rmW{{\mathbf{W}}}
\def\rmX{{\mathbf{X}}}

\DeclareMathAlphabet{\mathsfit}{\encodingdefault}{\sfdefault}{m}{sl}
\SetMathAlphabet{\mathsfit}{bold}{\encodingdefault}{\sfdefault}{bx}{n}

